# Advanced Image Processing for Astronomical Images

Diganta Misra[1], Sparsha Mishra[1] and Bhargav Appasani[1]

[1]School of Electronics Engineering, KIIT University, Bhubaneswar-751024, India

***Abstract***: Image Processing in Astronomy is a major field of research and involves a lot of techniques pertaining to improve analyzing the properties of the celestial objects or obtaining preliminary inference from the image data. In this paper, we provide a comprehensive case study of advanced image processing techniques applied to Astronomical Galaxy Images for improved analysis, accurate inferences and faster analysis.

***Keywords:*** Astronomy, Image Processing, Segmentation, Elliptical Galaxy.

## I. INTRODUCTION

Image Processing [1] is the collective term given to techniques or procedures used to process an image for analysis, feature extraction, object detection, et cetera. Image Processing has several applications in mostly all kind of domains including medical science, astronomy, automation industry amongst many others. With a huge volume of image data being generated or captured these days along with more powerful hardware including lenses and computational processing power, the popularity and necessity of Image Processing is increasing exponentially.

Image Processing along with Digital Signal Processing is highly important in Astronomy especially with the recent advancements in space exploration and the technological development of more robust and technically sound observatories with more powerful telescopes. The use of Image Processing and Digital Signal Processing in Astronomy [2] [3] varies from detection and classification or categorization of celestial objects, determining the distance from earth, understanding the physical properties of the subject in the image by performing spectrum analysis using the signal data.

With the recent advents in Machine Learning, astronomers and cosmological experts are having more tools at their disposal to understand our near celestial neighbours and Image Processing is undeniably one of the most crucial pre-processing and analytical steps in that pipeline. Currently, astronomers and cosmological scientists use the standard image processing and analysing systems for astronomy available which includes:

- AIPS (Astronomical Image Processing System) [4]: Originally designed using FORTRAN programming language by professionals at NRAO (National Radio Astronomy Observatory) in 1978, AIPS has been in use for 40 years now. AIPS provides a wide array of automated tools like Gaussian fitting of images, applying mathematical operators, spectra analysis, et cetera, for astronomers to analyze data considered in FITS (Flexible Image Transport System) format. Though being partially replaced by its to-be successor called CASA (Common Astronomy Software Applications) [5], formerly known as AIPS++, AIPS has evolved over the years and has received significant updates and remains popular to this date.

- IRAF (Image Reduction and Analysis Facility) [6]: IRAF, developed at NOAO (National Optical Astronomy Observatory), is an assemblage of software aimed at reducing astronomical images to their pixel array representation for advanced statistical analysis. IRAF is primarily confined to data obtained from imaging array detectors as CCDs (Charged Coupled Device). IRAF includes stacks of various applicative functionalities which includes determining redshifts of absorption or spectral analysis, the combination of images, calibration of fluxes and orientation of astronomical/ celestial objects captured within the image, compensation of variation in pixel sensitivity, et cetera.

Other Software based analyzing systems available include STSDAS (Space Telescope Science Data Analysis System), StarLink Project and many more. The existence of these automated frameworks have greatly improved the analytical pipeline and boosted research in astronomy in total.

## II. RELATED WORK

In [7], the Authors have written a book describing about imaging and manipulating images. It provides an in-depth analysis of how the image processing works. It helps people in learning about the incredible potential in digital imaging that has been unleashed by astronomy. In [8], The Authors have provided a description on an adaptive filter for processing for astronomical images which has been developed. The filter is capable is recognizing the local signal resolution and also adapts its own response to this resolution. The Authors in [9] have presented various methods that are used to measure the information in an astronomical image. The results achieved are targeted at information and relevance with a focus on experimental results in astronomical image and signal processing.

## III. IMAGE PROCESSING

Image Processing plays a vital role in understanding, analyzing and interpreting astronomical images. Starting from Image Smoothening, Noise removal, Edge Detection and Contour Mapping to Object Segmentation, digital image processing combined with signal processing is a powerful set of tools for astronomers to use while analyzing astronomical data. In the subsequent sub-sections, the research results along with the application of various mathematical algorithms and techniques have been described in detail.

## A. Extrema Analysis:

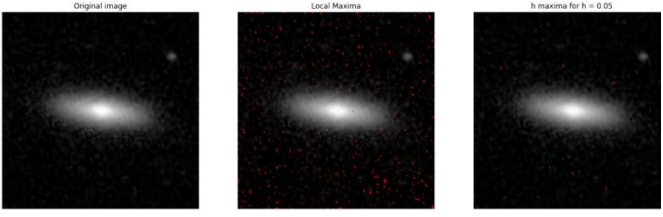

Fig. 1. (a). Original Elliptical Galaxy Image with label 806304. (b). Local Maxima of the Original Image. (c). h Maxima for h=0.05 of the Original Image.

Usually, in Galaxy Imaging, telescopes often capture images containing galaxies along with clusters of stars and other celestial objects. Correctly identifying the Galaxy within the image is the preliminary step before moving towards analyzing the galaxy subject. Extrema Analysis [10] proves to be extremely helpful in such cases to find regional maximas and minimas within the image for segmentation. Due to the noisy characteristics of the input image, $h$-maxima was applied with a magnitude of 0.05 for preferred results. With high level of noise, many local maximas were generated as seen in Fig. 1(b). The $h$ parameter is scaled with the dynamic range of the image and represents the grayscale level known as height by which the algorithm needs to descend to potentially reach a higher maximum which is technically local contrast observable in the image.

## B. Shape Index Analysis

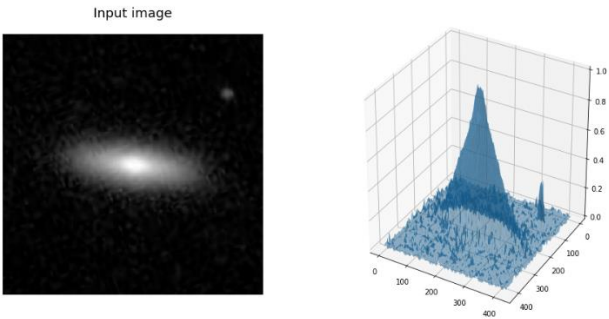

Fig. 2. (a). Original Input Image. (b). 3-dimensional visualization of shape index profile of the Input Image.

Preliminary analysis in astronomical image processing includes understanding the dimensional properties or the shape index profile of the celestial object in the image. Interpreting the shape index [11], orientation index and dimensional profile of the object helps astronomers to correctly identify the class it belongs to and also to conduct subsequent research on it. Shape Index Profile is a single-valued entity measuring the local curvature and is derived from the Eigen values of the Hessian. As seen in Fig. 2.(b)., the shape index [11] does get affected due to apparent noise pattern hampering the general texture of the image but is immune to uneven illumination.

Fig. 3 shows the shape index profile [11] of the input image along with spherical caps detection due to the $\sigma$ parameter being 1. This provides a clear intuition of the illumination concentration in the image, spatial orientation of the celestial object and also helps in defining the shape index of that object making it a crucial step in astronomical image analysis.

## C. Image Gradients

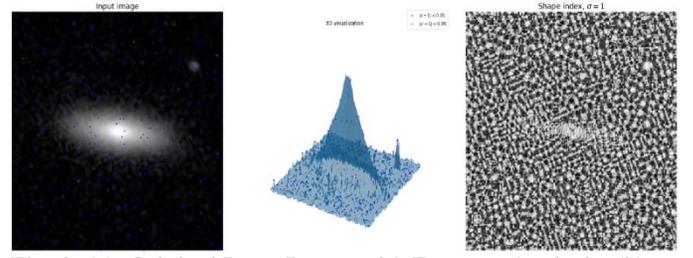

Fig. 3. (a). Original Input Image with Extrema Analysis. (b). 3-dimensional visualization of shape index profile of the Input Image. (c). Shape Index with $\sigma = 1$.

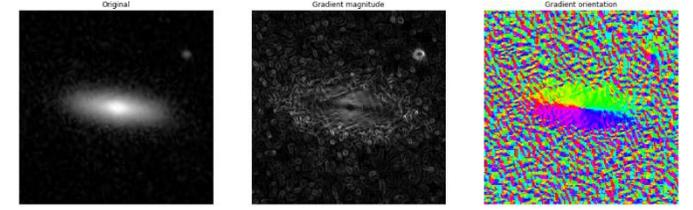

Fig. 4. (a). Original Image. (b). Gradient Magnitude of the Image. (c). Gradient Orientation in HSV colormap.

Image Gradients [12] are the fundamental building block of any digital image which represents a directional change in pixel intensity or contrast levels. Gradient Computation is a high priority task for many post-processing image processing techniques including edge detection and segmentation. For instance, Watershed Segmentation uses Local Gradients of the image to establish markers to define boundaries between objects in the image for Segmentation. Image Gradients computation is also used as a process for feature extraction and texture matching or pattern recognition within the image. Mathematically, Image Gradients are computed in the following way:

$$\nabla f = \begin{bmatrix} g_x \\ g_y \end{bmatrix} = \begin{bmatrix} \frac{\partial f}{\partial x} \\ \frac{\partial f}{\partial y} \end{bmatrix} \tag{1}$$

Basically, the gradients of an image can be defined to be the vector of its partial derivatives both in $x$-orientation and the $y$-orientation. In (1), $\frac{\partial f}{\partial x}$ is the gradient in the $x$-orientation and $\frac{\partial f}{\partial y}$ is the gradient in the $y$-orientation. These partial derivatives or individual gradients can be obtained by convolving a 1-dimensional filter to that image. The Gradient magnitude can subsequently be calculated by the following formula:

$$G = \sqrt{g_x^2 + g_y^2} \tag{2}$$

Lastly, the Gradient's direction can be obtained by deploying the following mathematical function which is represented as:

$$\theta = \tan^{-1} \begin{bmatrix} \frac{g_y}{g_x} \end{bmatrix} \tag{3}$$

$\theta$ is the angle of orientation of the gradients in the spatial domain.

Fig. 4 shows both the Gradient Magnitude Mapping and Gradient Orientation of the Input image of the elliptical galaxy. This provides a lot of information on the orientation of the object in the image and also is used in subsequent sections for image segmentation performed on the image to segment the galaxy from the image.

## D. Simple Cells Filter-Bank Analysis

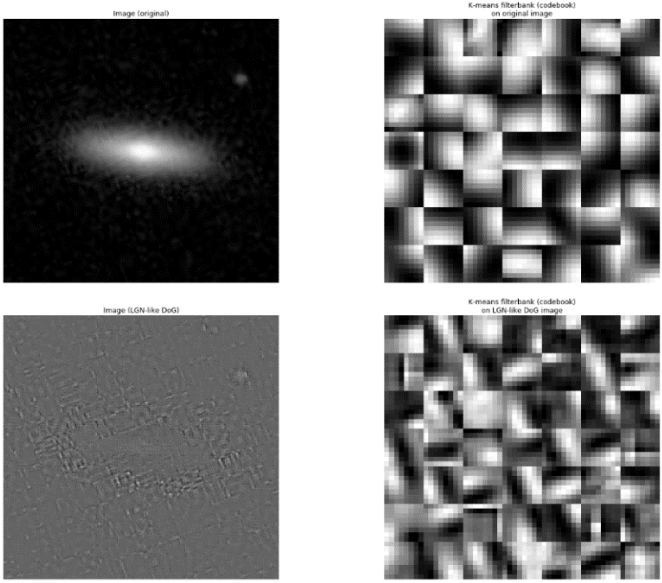

Fig. 5. (a). Original Image. (b). K-Means Filter-Bank of the Original Image. (c). Original Image constructed in perspective of Lateral Geniculate Nucleus (LGN) using Difference of Gaussians (DoG). (d). K-Means Filter-Bank on the LGN DoG constructed image.

Usually computing filter-banks involves heavy mathematical foundations for image classification, however, Simple cell analysis [13][14] is inspired by the receptive fields found in mammalian primary visual cortex by using simple Gabor filters on the retinal perspective of the original image as to construct the filter-bank as shown in Fig. 5 (a) and (b). Subsequently, the image was reconstructed in the perspective of Lateral Geniculate Nucleus (LGN) using Difference of Gaussians (DoG) approximation. Finally the filter-bank was computed on the LGN DoG generated image. To obtain the filter-bank K-Means algorithm was used as a biologically plausible simple Hebbian learning rule.

Gabor filters [15] are extensively used as primary low-level edge detection filters and can be defined to be a simple Gaussian kernel convoluted with a sine filter. In Convolutional Neural Networks, a deep learning approach towards Image analysis, Gaussian Gabor Filters /Kernels have been commonly used for low-level feature representation and understanding like smoothening and Edge Detection. In equation form, they can be represented as shown in equation (4):

$$Ga(x; \mu; \sigma) = \sin(x) * \frac{e^{\frac{-(x-u)^2}{2\sigma^2}}}{\sigma\sqrt{2\pi}} \qquad (4)$$

The Fourier transform of the impulse function of a Gabor Filter which is the sinusoidal wave multiplied to a Gaussian is the convolution of the Fourier transform of the Harmonic sinusoidal function and the Fourier transform of the Gaussian function. The filter thus has real and imaginary components representing orthogonal directions and can be mapped in a mathematical function as:

$$g(x, y; \lambda, \theta, \psi, \sigma, \gamma) = e^{-x'^2 + \frac{\gamma^2 y'^2}{2\sigma^2}} e^{i\left(\frac{2\pi x'}{\lambda} + \psi\right)} \qquad (5)$$

$$g(x, y; \lambda, \theta, \psi, \sigma, \gamma) = e^{-x'^2 + \frac{\gamma^2 y'^2}{2\sigma^2}} \cos\left(\frac{2\pi x'}{\lambda} + \varphi\right) \qquad (6)$$

$$g(x, y; \lambda, \theta, \psi, \sigma, \gamma) = e^{-x'^2 + \frac{\gamma^2 y'^2}{2\sigma^2}} \sin\left(\frac{2\pi x'}{\lambda} + \varphi\right) \qquad (7)$$

Where x´ and y´ are represented using the formulas shown in the equation (8) and (9).

$$x' = x\cos\theta + y\sin\theta \qquad (8)$$

$$y' = -x\sin\theta + y\cos\theta \qquad (9)$$

In the equations (5), (6) and (7); λ represents the wavelength of the sinusoidal factor, φ represents the phase offset and γ is the spatial aspect ratio and it specifies the ellipticity of the support of the Gabor function.

Gaussian filters have often been labeled to be the closest approximation of human vision level of perception of how our visual cortex understands the underlying patterns in any environment that it visually perceives. 2d Gaussian filters amplified with a desired frequency can be very useful in performing feature extraction on an image. 2-D Gaussian filters can be represented in a discrete domain as follows:

$$G_c[i, j] = B_e^{\frac{-(i^2 + j^2)}{2\sigma^2}} \cos(2\pi f(i\cos\theta + j\sin\theta)) \qquad (10)$$

$$G_s[i, j] = C_e^{\frac{-(i^2 + j^2)}{2\sigma^2}} \sin(2\pi f(i\cos\theta + j\sin\theta)) \qquad (11)$$

Here, B and C are the normalizing factors to be estimated, f is the frequency which is being looked for in the texture and by varying Θ, we can look for texture oriented in a particular vector direction. By varying σ, which is the characteristic standard deviation, we can modulate the size or the area of the image to be analyzed.

A normal Gaussian distribution is a peak-shaped function over a range of values defined by x, it's mean μ and the standard deviation to be σ as shown in equation (12):

$$G(x; \mu; \sigma) = \frac{e^{\frac{-(x-u)^2}{2\sigma^2}}}{\sigma\sqrt{2\pi}} \qquad (12)$$

Fig. 6 shows the visualization of a Gaussian Kernel and a Gabor Kernel respectively.

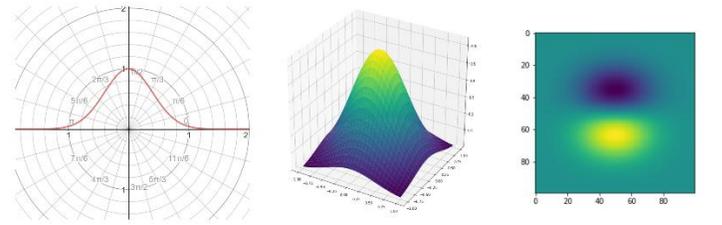

Fig. 6. (a). Gaussian 2-dimensional representation. (b). Gaussian 3-dimensional representation. (c). 2-dimensional Gabor Filter

Difference of Gaussian (DoG) [16], a very similar approximation of Laplacian of Gaussian (LoG) takes the difference between≈ two Gaussian Smoothened Images where the blobs are detected from the scale-space extrema of the difference of Gaussians. Mathematically, the DoG algorithm can be represented as:

$$\nabla^2_{norm} L(x,y;t) \approx \frac{t}{\Delta t}\big(L(x,y;t+\Delta t) - L(x,y;t)\big) \quad (13)$$

where $\nabla^2 L(x,y;t)$ is the Laplacian of the Gaussian Operator defined in Laplacian of Gaussian (LoG) to be:

$$\nabla^2 L = L_{xx} + L_{yy} \quad (14)$$

Difference of Gaussian (DoG) is primarily used for blob detection on an image but here has been incorporated as an approximation algorithm on the LGN generated image.

LGN (Lateral Geniculate Nucleus) is an active relay region in the thalamus for the visual pathway. It is the focal point in the visual cortex and perceives sensory data obtained from the retina. It's made up of neuron layers and optic fibers, and connects the optic nerve to occipital lobe.

In Fig. 5, the filter-bank constructed represents the simple cells in perspective to primary cortex input sensory data. These filter banks can be represented to be simple edge detector kernels applied on to the image.

### E. Non-Local Means Noise Removal

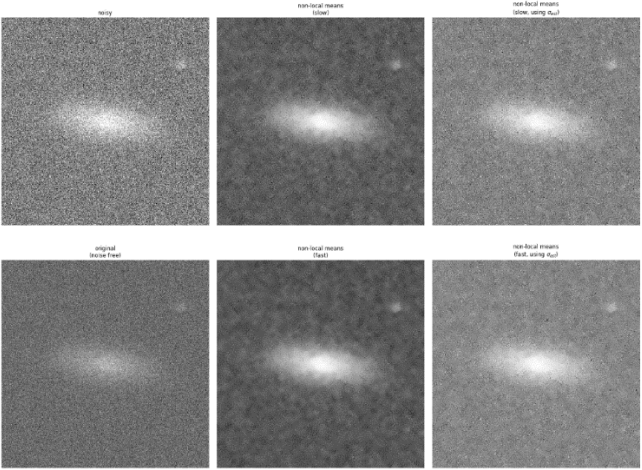

Fig. 7. Non-Local Means for De-noising for both amplified noisy input image and standard noisy image.

Noise removal remains an important task in image processing pipeline to smoothen the image and to maintain the original information within the image. Conventional noise removal methods do have a trade-off while being successful in removing noise by smoothening the image, they often tend to fail to preserve the edges present in the image which are highly important in the post-processing tasks. While dealing with astronomical images, it's necessary to preserve edges and remove noise simultaneously while maintaining the original texture patterns of the image.

As shown in Fig. 7, Non-Local Means filter [17] was applied on the noisy image having estimated noise standard deviation to be 0.35677915437197705. The non-local means algorithm follows the procedure of replacing the intensity value of the target pixel with average of a selection of intensities of other pixels where small regions centered on other pixel is compared to the region having the target pixel as its center and the averaging is performed when both the regions have a high rate of similarity thus helping in preserving details and texture present in the image. In the analysis, both fast non-local filter and slow non-local means using $\sigma_{est} = 0.3567791543719770$, which is the estimated noise standard deviation was applied. During fast non-local denoising, uniform spatial weighting is applied on the regions whereas when using slow non-local means, a spatial Gaussian Weighting is applied on the regions for computing the distance or the similarity index between them.

### F. Self-Tuned Restoration of Image

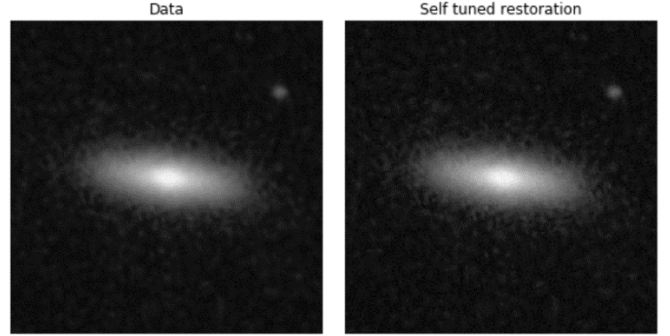

Fig. 8. (a). Original Input Image. (b). Self-tuned Restored Image using Weiner and Unsupervised Weiner Filter.

Non-Linear methods of removing noise in an image may be highly effective in preserving the sharp edges present in the image while removing the noise pattern in the image but has a major trade-off in the form of requiring more computational power and being slower. Weiner and Unsupervised Weiner Algorithms are Linear Models and hence are considerably faster although they fail in preserving sharp detail edges present in the image but are highly efficient in smoothening the image and removing noise. In Fig. 8, the original image was de-convolved with a Weiner and Unsupervised Weiner Filter.

Weiner De-convolution [18] is a popular process of noise removal in digital images in the frequency domain. Weiner filter is based on PSF (Point Spread Function), the prior regularization applied (penalization of high frequency) and the balancing trade-off between the data and prior adequacy. The Unsupervised Weiner Filter is based on an iterative Gibbs Sampler having a self-tuned regularization parameter based on data learning, which draws alternative sampling of posterior conditional law of the image, the noise power domain of the image and the image frequency power.

Mathematically, Weiner De-convolution can be defined as shown in the following equations:

$$y(t) = (h*x)(t) + n(t) \quad (15)$$

$y(t)$ is the system represented by the summation of $n(t)$, the noise signal with the convolution of $h(t)$, which is the impulse response of the linear time-invariant (LTI) system and $x(t)$ which is the original signal at any time $t$. The Weiner De-convolution provides an appropriate solution $g(t)$ to the following equation to minimize mean squared error.

$$\hat{x}(t) = (g*y)(t) \quad (16)$$

$\hat{x}(t)$ is the estimate of $x(t)$. Weiner de-convolution can thus be represented in the frequency domain to be:

$$G(f) = \frac{H^*(f)S(f)}{|H(f)|^2 S(f) + N(f)} \tag{17}$$

$G(f)$ and $H(f)$ represent the Fourier transforms of g and h respectively at frequency $f$. $S(f)$ is defined to be the mean power spectral density of the original signal $x(t)$. $N(f)$ is the mean power spectral density of the noise signal $n(t)$ and $H^*(f)$ is the complex conjugate representation. The obtained $G(f)$ can be then applied to (16) in the frequency domain which will give $X(f)$ as output which can be converted to the de-convoluted signal $x(t)$ by performing inverse Fourier transform on it.

Due to faster performance and efficient noise removal, self-tuned restoration can be used to accelerate processing of astronomical images for faster analysis as shown in Fig. 8.

### G. Chan Vese Segmentation

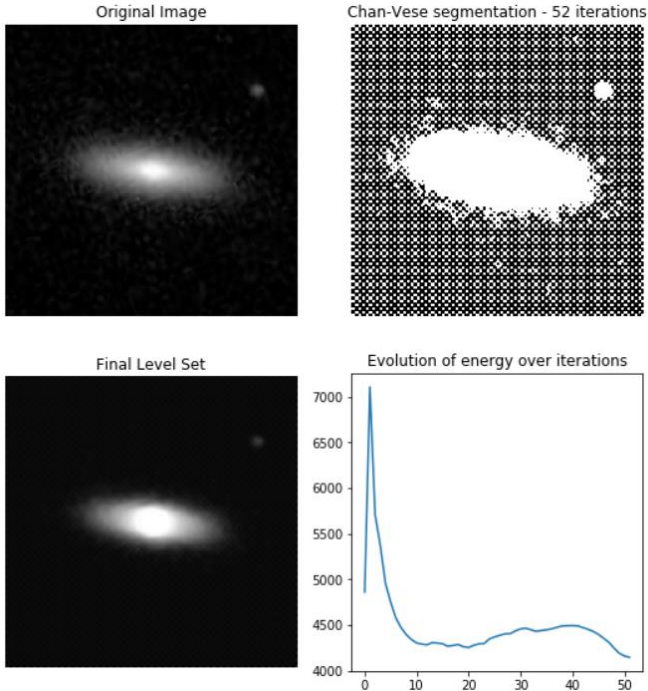

Fig. 9. Chan Vese Segmentation over 52 iterations with the evolution of energy over 52 iterations.

Object Segmentation remains a crucial step in all kinds of image processing and computer vision problem statements. Chan Vese Segmentation [19][20][21] involves an algorithm used for segmenting objects lacking definitive boundaries. Most Astronomical Images obtained are noisy and grayscale and lack definitive boundary confining the celestial object in the image. Chan Vese Segmentation, rather than using active contour modelling based on edge detection and sharp contrast variations, is based on level sets which evolve over iterations to reduce the energy to a minimum which is defined by defined by weighted values corresponding to the sum of difference in intensities from the average value outside the region segmented, the sum of the difference from the average values within the segmented region and an unique term which has a dependency on the length of the boundary of the segmented region.

The Chan Vese algorithm involves a certain list of parameters including $\mu$ which usually ranges between 0 and 1

and here was kept to be 0.5, $\lambda_1$ and $\lambda_2$ which are usually kept to be 1 but due to the irregular distribution of the objects with respect to the background their values were kept to be 1 and 2 respectively with a maximum iteration value of 200. As we see from Fig. 9, the desired result was achieved in 52 iterations. This can be extremely useful in celestial object segmentation and identification in astronomical images.

### H. Random Walker Segmentation

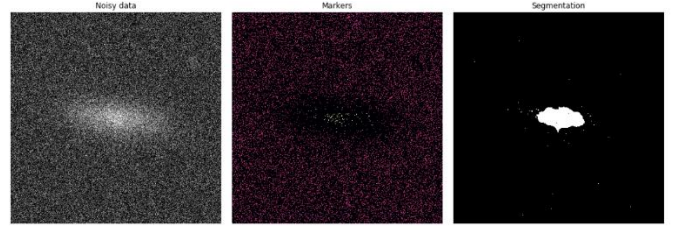

Fig. 10. (a). Original Input Image amplified with Salt and Pepper Noise. (b). Markers computed on the noisy image. (c). Segmented Image.

As discussed in the previous section, segmentation remains a high priority task, it also involves in scenario while dealing with noisy input data which is common in case of astronomical images. Here, Random Walker Algorithm was used for Segmentation of the original image modulated with synthetic salt and pepper noise.

Random Walker Algorithm [22] involves a set of markers responsible for labelling the phases present in the image which can be anything from 2 or above. The algorithm involves an anisotropic diffusion equation solved using these labels initiated at the markers' positions where the local diffusivity co-efficient is greater if neighboring pixels have similar intensity values and thus making diffusion difficult across the high gradients. Here, Random Walker algorithm was initiated using the tail end values of the Histogram of gray values obtained from the noisy image. As seen in Fig. 10, it is highly effective in segmenting the elliptical galaxy within the noisy image proving to be highly effective in object segmentation tasks in astronomical image processing.

### I. Power Spectrum Analysis

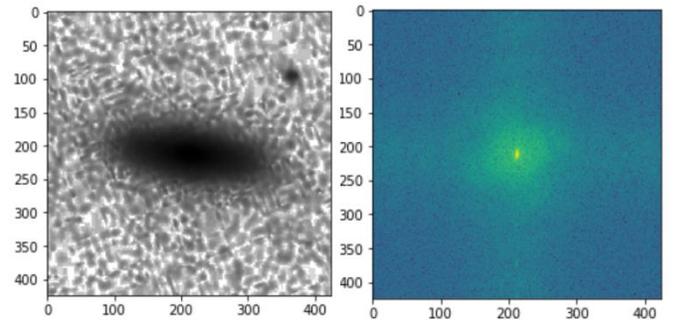

Fig. 11. (a). Grayscale inverted original galaxy image. (b). 2-dimensional Power Spectrum.

Images are a 2-dimensional form of a signal and within Signal Analysis, Spectrum Analysis is one of the most important processes to understand the signal and its subsequent properties. Applying Fourier Transforms to astronomical images have many varied applications from noise removal, finding small structures in diffused galaxies, et cetera.

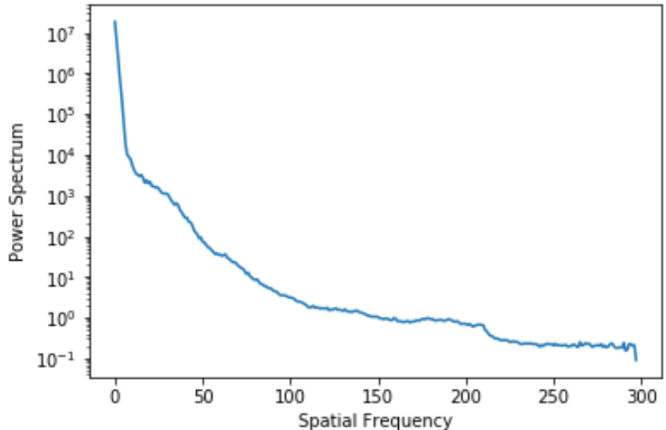

Fig. 12. Azimuthally averaged 1-dimensional Power Spectrum of the input galaxy image.

Power Spectrum [23] is a powerful signal analysis method which involves plotting the portion of power of the signal within the given range of frequency. This can be obtained by applying inverse Fourier Transform to the given signal. Power Spectral density gives the intuition on the dominant frequencies within the image which are extremely helpful in post-processing analysis like edge detection, contour modelling, compression of the image, et cetera. Fig 11 and 12 show the power spectra both in 2-D and 1-D with respect to the spatial frequency range of the image data provided.

*J. Overlapping Distance Mapping using Watershed Segmentation*

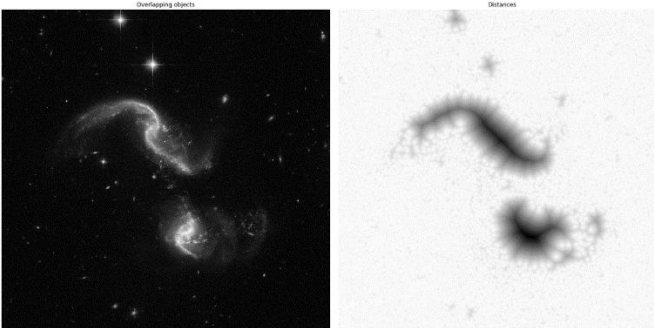

Fig. 13. (a). Grayscale inverted original galaxy image named "Crash in Progress" (ESA/Hubble, NASA) (b). Distances computed using Watershed Segmentation.

Usually, astronomical telescopes capture images having overlapping celestial objects primarily galaxies or same within extremely close proximity. Measuring the distance between the two becomes crucial in understanding their related dimensional properties. As shown in Fig. 13, Watershed Segmentation, a popular Segmentation algorithm was used to plot the representation of the distance mapping between the two galaxies of the Apr 256 system captured by Hubble's Advanced Camera for Surveys (ACS) and the Wide Field Camera 3 (WFC3) released in 2018 with the system stationed at a distance of 350 million light-years away.

Watershed Algorithm [24] is a classic segmentation algorithm used in Image Processing. It follows a strict procedure of segmenting the image based on the markers obtained which are computed based on the area of low gradient value in the image. Technically, an area of high gradient in the image defines the boundaries separating the objects present in the image. This proves to be extremely helpful in defining distances between overlapping celestial objects within an astronomical image.

## IV. Experimental Set-up

The research was conducted using data obtained from Sloan Digital Sky Survey and ESA/Hubble, NASA. All the software simulations were conducted using Python programming language along with its sub-modules and packages including: Scikit-Image, OpenCV, Matplotlib, Seaborn, PyLab, Scipy and Numpy on a dedicated Jupyter Notebook server. The hardware specifications of the system used are as follows: MSI GP-63 8RE Leonard equipped with Intel core i7-8[th] gen processor, NVIDIA GTX 1060 GPU on a Windows 10 Professional Operating system.

## V. Conclusion

The research is aimed to provide academia and astronomers a concrete comprehensive guide towards performing image processing on astronomical images. It also defines the benchmark of the performance of algorithms capable of being deployed on astronomical images and can be used as a reference for future research in improving the analytical pipeline of astronomical image processing. Future work includes defining and constructing an automated software pipeline efficient enough to provide analytical and statistical results based on a given input astronomical image.